\documentclass{article}
\usepackage[preprint]{colm2026_conference}

\usepackage{microtype}
\usepackage{hyperref}
\usepackage{url}
\usepackage{booktabs}
\usepackage{graphicx}
\usepackage{enumitem}
\usepackage{multirow}
\usepackage{wrapfig}

\usepackage{lineno}

\definecolor{darkblue}{rgb}{0, 0, 0.5}
\hypersetup{colorlinks=true, citecolor=darkblue, linkcolor=darkblue, urlcolor=darkblue}

\title{Rate--Utility Frontiers for Language Encodings:\\
Comparing Tokens, Bytes, and Pixels Under\\Controlled Linguistic Content}

\author{%
Ingo Ziegler \quad Martin Krebs \quad Desmond Elliott\\
Department of Computer Science, University of Copenhagen\\
\texttt{inzi@di.ku.dk}, \texttt{lsw275@alumni.ku.dk}, \texttt{de@di.ku.dk}
}

\begin{document}

\ifcolmsubmission
\linenumbers
\fi

\maketitle

\begin{abstract}
Language models encode text as subword tokens, raw bytes, or rendered pixels, but these encodings are usually compared under modeling constraints that expose different amounts of linguistic content to models across different languages.
We instead ask what each encoding preserves when both the content and the downstream capacity are controlled.
Using verified parallel sentences across thirteen languages and five scripts, we compare tokens, bytes, and pixels through a shared bottleneck whose width is swept to trace rate--utility frontiers.
This separates three quantities that are often conflated:
the number of input positions an encoding creates,
the latent capacity available after encoding,
and the task-relevant information that survives compression.
We evaluate three utilities: surface form preservation, cross-lingual sentence alignment, and topic classification.
No encoding dominates across tasks or capacity regimes.
Pixels preserve surface form best, bytes preserve cross-lingual alignment best, especially in same-script multilingual settings, and tokens support topic prediction best.
These performances are not explained by sequence length alone.
Short inputs can discard useful meaning, while long inputs can preserve information that compresses well.
Choosing an encoding is therefore not a fixed preference for tokens, bytes, or pixels, but a rate--utility tradeoff that depends on the task, language mix, capacity regime, and compute budget.
\end{abstract}

\section{Introduction}
Before a language model can process text, the text must be encoded as a sequence of units.
Most models use subword tokens~\citep{radford2019language,grattafiori2024llama3herdmodels,yang2025qwen3technicalreport,glm5team2026glm5vibecodingagentic}, some read raw bytes~\citep{xue-etal-2022-byt5,yu2023megabyte,pagnoni-etal-2025-byte}, and a few render text as images and process pixels~\citep{salesky-etal-2021-robust,rust2023pixel,kesen-etal-2025-multilingual}.
Although these encodings preserve linguistic content, they change the representation the model sees.
Tokens form short sequences of meaning-bearing units from a learned but fixed vocabulary~\citep{sennrich-etal-2016-neural}.
Bytes form longer sequences from a small, universal alphabet~\citep{xue-etal-2022-byt5}.
Pixels form a two-dimensional image, with spatial layout and visual detail but no discrete symbols~\citep{rust2023pixel}.

\begin{figure}[t]
\centering
\includegraphics[width=\linewidth]{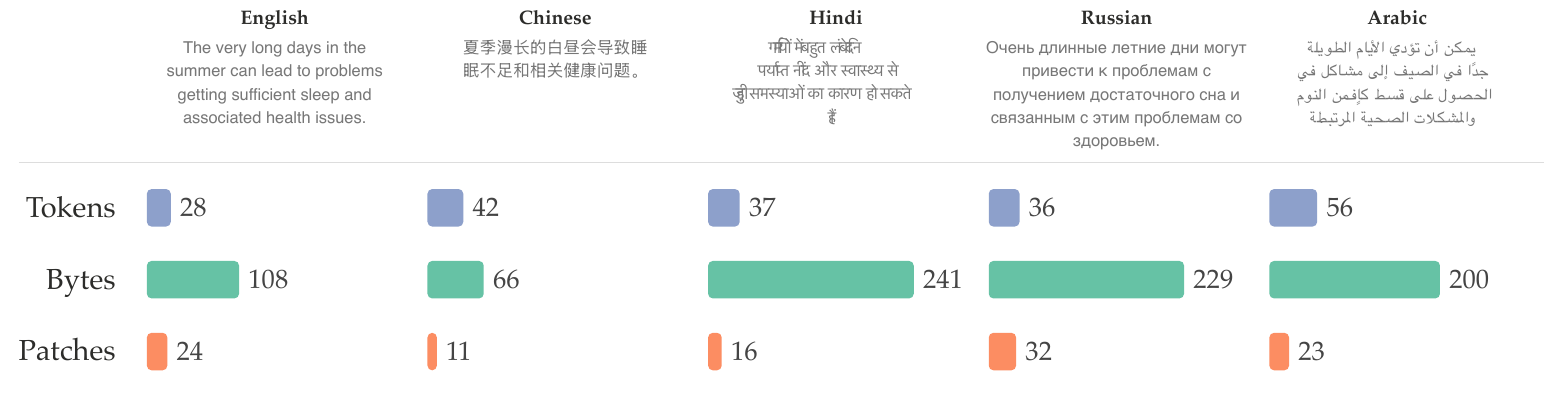}
\caption{%
The same linguistic content produces different \emph{source rates} across encodings and languages.
Each column shows a translation of the same SIB-200 sentence, with token, byte, and pixel patch lengths annotated.
A fixed input length does not expose equal content across representations to a model.
For example, Chinese is longer than English in tokens, even with tokenizers re-trained on matched-content and regime specific languages, but shorter in bytes and patches.
Our experiments examine how source rate differences interact with bottleneck capacity across tasks that require different kinds of information.
}
\label{fig:teaser}
\end{figure}

These differences widen across languages and scripts, which makes encodings hard to compare.
Prior work has shown that pre-trained tokenizers expose unequal amounts of linguistic content across languages~\citep{petrov2023language,ahia-etal-2023-languages}.
Figure~\ref{fig:teaser} shows that the problem generalizes beyond tokenization.
A fixed sequence length, batch size, or compute budget can hold different amounts of content depending on both the encoding and the language~\citep{petrov2023language}.
A difference measured between two encodings may therefore reflect how much content the model was shown, or how much capacity it was given, rather than what the encoding preserves.

A controlled comparison needs two quantities held constant.
First, the content must be identical, so that nothing but the encoding varies.
Second, the downstream capacity must be controlled, so that an encoding is judged by how well it uses a given budget, not by how large a budget it takes.
This changes the central question from which encoding is shorter to which encoding preserves useful task-relevant information under compression.

We meet both conditions directly.
We work with SIB-200~\citep{adelani-etal-2024-sib}, a human-translated parallel dataset, where the same meaning appears in every language, and we pass every encoding through a shared bottleneck of fixed width.
Complementary to~\citet{ahia-etal-2023-languages} and~\citet{petrov2023language}, we re-train tokenizers on matched content for each language regime, so the token baseline is adapted to the same languages as the byte and pixel encodings.
This lets us separate three quantities that are often collapsed:
how many input positions an encoding creates,
how much latent capacity we allow after encoding,
and which task-relevant information survives compression.
We call the first the \emph{source rate}, sweep the second by varying the bottleneck width, and measure the third through three utilities that probe complementary uses of text representations:
preserving form, aligning translations, and supporting downstream semantic prediction.
We evaluate these quantities across language regimes that range from a single language, to five languages in one script, to five scripts at once.

We find that the utility frontiers do not follow source rate.
Instead, encoding comparisons depend strongly on how content and capacity are controlled.
A short encoding can misallocate capacity, and a long encoding can preserve structure that compresses well.
The preferred encoding changes with the task and the capacity regime: pixels preserve surface form best, bytes preserve cross-lingual alignment most reliably, and tokens support topic prediction best.
The same script can also change roles across utilities.
Chinese is difficult to align across scripts, especially for pixels, yet its logographic form narrows the gap between encodings for topic prediction.

Our results demonstrate that a cheaper encoding is not necessarily a better one, and what counts as useful depends on what one wants to model.
Our contributions are as follows:
\begin{itemize}
\item We introduce a controlled rate--utility comparison for text encodings.
The comparison holds linguistic content fixed with parallel sentences, fixes downstream capacity with a shared bottleneck, and traces utility as capacity changes, together with the training compute each frontier requires.
\item We show that encoding preferences are utility-dependent.
Pixels preserve surface form, bytes preserve cross-lingual alignment, and tokens support topic prediction, with script mix and bottleneck capacity changing the margins.
\item We measure source rate for tokens, bytes, and pixels on identical linguistic content across thirteen languages and five scripts, confirming and extending prior findings on length disparities to rendered text and matched-content tokenizer regimes.
\end{itemize}

\section{Related Work}

\paragraph{Tokenization and multilingual imbalance.}
Prior work has shown that multilingual tokenizers differ in fertility~\citep{nayeem2025beyond}, vocabulary coverage~\citep{limisiewicz-etal-2023-tokenization}, and downstream quality across languages~\citep{rust-etal-2021-good}, especially for low-resource~\citep{dewangan-etal-2025-every} and morphologically rich languages~\citep{asgari2025morphbpemorphoawaretokenizerbridging} and for scripts underrepresented in the tokenizer training data~\citep{limisiewicz-etal-2023-tokenization}.
Closest to our source-rate analysis,~\citet{petrov2023language} and~\citet{ahia-etal-2023-languages} measure length disparities on parallel text from FLORES-200~\citep{nllbteam2022languageleftbehindscaling}.
They show that the same content can cost more tokens in one language than another, with~\citet{petrov2023language} also reporting similar gaps at the byte and character level.
These studies focus on pre-trained tokenizers and frame length disparity as a question of fairness, access, and commercial cost.
We re-train a tokenizer per language regime on matched parallel text, which isolates source rate also from the tokenizer's own training data, and we add rendered pixels as a third encoding.
Centrally, we move past length itself by asking which task-relevant information each encoding preserves once every encoding passes through the same capacity bottleneck.

\paragraph{Vocabulary-free encoders.}
These replace learned subword vocabularies with smaller atomic units such as UTF-8 bytes or Unicode characters.
ByT5~\citep{xue-etal-2022-byt5} uses uncompressed bytes, while CANINE~\citep{clark-etal-2022-canine} downsamples characters before its deeper layers.
Pixel models~\citep{salesky-etal-2021-robust,rust2023pixel} drop symbols entirely by rendering text to images and processing pixel patches, and Pix2Struct~\citep{lee2023pix2struct} reads language directly from rendered screenshots.
Both pay a structural cost: bytes in sequence length and pixels in a two-dimensional image.
These works ask whether such models can match a token model after changing the architecture, objective, or scale.
We ask instead what each encoding preserves when architecture, content, and capacity are held fixed.

\section{Experimental Setup}
\label{sec:setup}

\subsection{Controlled content and language regimes}
\label{sec:setup_content}
Our comparison controls for content, so that any differences we measure come from the encoding, not from the content shown to the model.
Therefore, all experiments use SIB-200~\citep{adelani-etal-2024-sib}, a topic-labeled, human-translated parallel dataset.
Each sentence appears in every language with the same topic label, so content and labels are aligned across languages by design.
We use the standard splits: 701 training, 99 validation, and 204 test sentences per language.

We study five language regimes that isolate one factor at a time.
Two are monolingual: \textbf{English}, written in the alphabetic Latin script, and \textbf{Chinese}, written in the logographic Han script.
Two are multilingual within a single script: \textbf{Latin-5} (English, Spanish, Turkish, Indonesian, and Swahili) and \textbf{Cyrillic-5} (Russian, Bulgarian, Ukrainian, Serbian, and Kazakh).
The fifth, \textbf{Multiscript-5}, pairs five languages with five scripts: English in Latin, Russian in Cyrillic, Arabic in Arabic, Hindi in Devanagari, and Chinese in Han.
Each multilingual regime holds the same sentences as a monolingual one, repeated once per language, so it adds languages but no new content.
The monolingual regimes remove multilinguality, the same-script regimes add languages while holding the script fixed, and the multiscript regime adds scripts.
Considering each independently separates the effect of language count from the effect of script diversity.

We use three probes, each corresponding to a different use of text representations, to measure which information each encoding preserves under compression.
Form preservation asks whether the written sentence itself can be recovered.
Cross-lingual retrieval asks whether translations of the same sentence stay close in representation space.
Topic classification asks whether the compressed representation retains enough information to support a downstream prediction head.
Each task is defined in full where it is used in Section~\ref{sec:results}.

These probes matter for different applications.
Form matters when the surface itself is part of the task, as in OCR~\citep{kim2022ocr,li2023trocr}, scene text~\citep{Jang_2026_WACV}, document understanding~\citep{lee2023pix2struct,kim2022ocr}, spelling-sensitive processing~\citep{salesky-etal-2021-robust,vesalainen2026errorpatternshistoricalocr}, and visual text generation~\citep{liu-etal-2023-character,liu2024glyph}.
Cross-lingual meaning matters when representations require semantic equivalence across scripts and languages, such as bitext mining~\citep{heffernan-etal-2022-bitext,artetxe-schwenk-2019-massively}, multilingual sentence embedding~\citep{reimers-gurevych-2020-making,enevoldsen2025mmteb}, or cross-lingual retrieval~\citep{zhang-etal-2023-miracl}.
Topic classification captures the setting closest to encoder-style inference~\citep{wang-etal-2018-glue,devlin-etal-2019-bert}, where a fixed representation must support a classification head.

\subsection{Encodings}
\label{sec:setup_encodings}
We compare three encodings of the same content.
Tokens expose a learned segmentation: frequent subwords become single units, which compresses common text but fixes a vocabulary.
Bytes expose a fixed universal alphabet: every string is a sequence of UTF-8 bytes, at the cost of more positions.
Pixels expose visual form: text is rendered to an image, which removes symbolic units but adds surface detail such as spacing and glyph shape.

\textbf{Tokens.} We use SentencePiece~\citep{kudo-richardson-2018-sentencepiece} with a Unigram model~\citep{kudo-2018-subword} and a vocabulary of 32{,}000.
We train one tokenizer per language regime on the union of Bible translations~\citep{christodoulopoulos2015massively,wordproject_bible,htmlbible_ukrainian} for that regime's languages, then apply it to SIB-200.
As the Bible translations are parallel, tokenizer training is matched across languages within each regime.
The shift from Bible text to SIB-200 leaves some vocabulary slots unused due to the small dataset size.

\textbf{Bytes.} We encode text as raw UTF-8 bytes, a fixed alphabet of 256 values.

\textbf{Pixels.} We render text in grayscale with the size 16 Noto Sans fonts onto a grid with maximum height of 36 pixels.
We slice each rendered line into non-overlapping patches of $36\times32$ pixels, so each patch is 32 pixels wide, and treat a sentence as a one-dimensional sequence of patches~\citep{rust2023pixel}.
We render right-to-left scripts in their natural order and never truncate across any modality.
Shorter lines are padded to the regime's maximum sequence length.
Additional details on encodings can be found in Appendix~\ref{app:encoding-details}.

\subsection{Encoder model and capacity bottleneck}
\label{sec:setup_model}
All experiments share the same Transformer-based~\citep{vaswani2017attention} encoder architecture, so that a difference between encodings reflects the encoding and not the model.
Only the part attached after the bottleneck changes across experiments, and we describe those parts where they are used.

Each encoding is first mapped to its embedding.
Tokens and bytes use an embedding table.
Pixels use a linear projection of each $36\times32$ patch, equivalent to a non-overlapping convolution~\citep{rust2023pixel}.
The resulting sequence passes through a standard pre-norm Transformer~\citep{pmlr-v119-xiong20b} encoder.
Across all experiments and encodings, total model sizes vary between 5.6M and 13.8M parameters.
Full hyperparameters and additional implementation details are provided in Appendices~\ref{app:encoding-details}--~\ref{app:task-objectives}, and we provide code for all experiments at \href{https://github.com/ziegler-ingo/rate-utility-frontiers}{github.com/ziegler-ingo/rate-utility-frontiers}.

We then compress the encoder output through a bottleneck.
The bottleneck is a single learned query that attends to the encoder output and returns one vector, in the style of a Perceiver~\citep{pmlr-v139-jaegle21a}.
This holds downstream representational capacity fixed across encodings.
Without it, an encoding that exposes more positions could carry more total information, breaking the controlled comparison.
The width of this vector, $D$, is the capacity available to every downstream task: all information must pass through it.
We treat $D$ as the controlled variable and sweep it across fifteen values, from 256 down to 1: $\{256, 128, 64, 48, 32, 24, 20, 16, 12, 10, 8, 6, 4, 2, 1\}$.
A smaller $D$ forces more compression.
Plotting utility against $D$ gives a frontier:
one encoding is more compact than another for a given utility if it reaches the same score at a smaller $D$, or a higher score at the same $D$.

This sweep is important because source length and preserved information need not move together.
An encoding can expose many positions but store the information needed for a task in a narrow latent.
Another encoding can expose fewer positions but require more latent capacity to preserve the same utility.
A single bottleneck width would hide this distinction.
The frontier shows how each encoding degrades as capacity is removed, and therefore measures not only whether an encoding works, but how compactly it stores the information a task needs.

\section{Results}
\label{sec:results}

We report results in three parts.
We first measure the source rate of each encoding, a fixed property of the data.
We then train models and trace how much utility each encoding preserves as we vary the bottleneck width $D$, for form (Section~\ref{sec:results_form}), alignment (Section~\ref{sec:results_meaning}), and prediction (Section~\ref{sec:results_predictive}), and then measure what each frontier costs to train (Section~\ref{sec:results_flops}).
We discuss limitations and future work in Appendix~\ref{app:limitations-and-future-work}.

\begin{figure}[t]
\centering
\includegraphics[width=\linewidth]{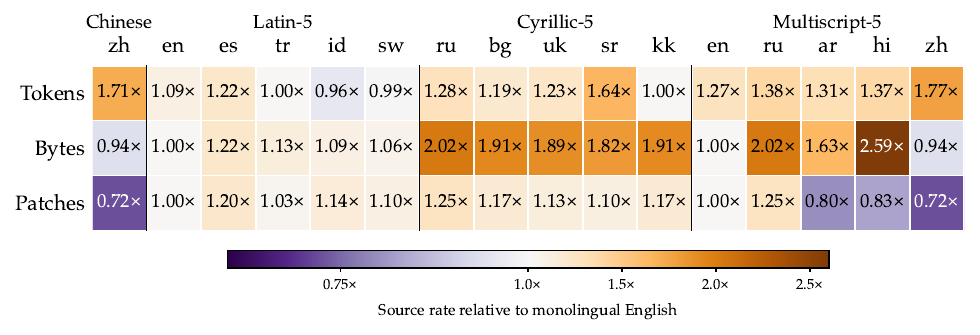}
\caption{%
Source rate relative to monolingual English, by language and encoding.
Each cell is a language's mean source rate divided by English's mean in the same encoding.
Values above $1\times$ are more expensive than English, below $1\times$ are cheaper.
The cheapest encoding depends on the language: tokens are cheapest for English yet most expensive for Chinese, while pixel patches are cheapest for Chinese, Arabic, and Hindi.
}
\label{fig:heatmap}
\end{figure}

\subsection{Source rates}
\label{sec:results_source}

\begin{wraptable}{r}{0.38\linewidth}
\vspace{-1em}
\centering
\small
\setlength{\tabcolsep}{2pt}
\begin{tabular}{lrrr}
\toprule
Regime & Token & Byte & Patch \\
\midrule
English       & 42 & 129 & 29 \\
Chinese       & 71 & 120 & 21 \\
Latin-5       & 44 & 142 & 31 \\
Cyrillic-5    & 53 & 246 & 33 \\
Multiscript-5 & 59 & 210 & 26 \\
\bottomrule
\end{tabular}
\vspace{-1em}
\caption{Mean units per sentence.}
\label{tab:source_means}
\vspace{-1em}
\end{wraptable}
The source rate is the number of positions an encoding produces for a piece of content: tokens after segmentation, bytes, or image patches after rendering.
It varies across encodings and languages even when the content is held constant~\citep{petrov2023language,ahia-etal-2023-languages}.
We confirm their findings in our controlled setup, extend them to rendered pixels and to tokenizers trained per regime.
We then ask whether these source-rate differences translate into utility differences in the following experiments.
We measure on all 1004 SIB-200~\citep{adelani-etal-2024-sib} sentences per language, averaged within each regime.
In raw sequence length the three encodings are ordered consistently: patches are the fewest positions, tokens the middle, and bytes the most (Table~\ref{tab:source_means}).

The cheapest encoding depends on the language.
We make this visible by normalizing each language to monolingual English within each encoding (Figure~\ref{fig:heatmap}).
Chinese is the clearest case.
A Chinese sentence costs $1.71\times$ as many tokens as English, but only $0.94\times$ as many bytes and $0.72\times$ as many patches.
The ranking inverts: tokens are the most expensive encoding for Chinese and the cheapest for English.
The tokenizer, even when trained only on matched-content Chinese, segments a Chinese sentence into more units than an English one.
A Han character is three bytes but a whole morpheme, so it needs fewer positions and less visual space than meaning-aligned alphabetic versions.
The pattern holds across scripts: dense non-Latin scripts are costly in bytes, with Hindi highest at $2.59\times$ English, yet the same scripts are the cheapest in pixels, with Chinese, Arabic, and Hindi all below English ($0.72$--$0.83\times$ the patches).

These source rates show that the cheapest encoding depends on the language and the script mix, so the standard way of comparing encodings is confounded.
A fixed sequence length or compute budget therefore exposes each encoding to a different amount of content~\citep{petrov2023language,ahia-etal-2023-languages}, and which language is disadvantaged changes with the encoding.
The rest of the paper removes this confound by holding content fixed and varying only the capacity each encoding is allowed.

\begin{figure}[t]
\centering
\includegraphics[width=\linewidth]{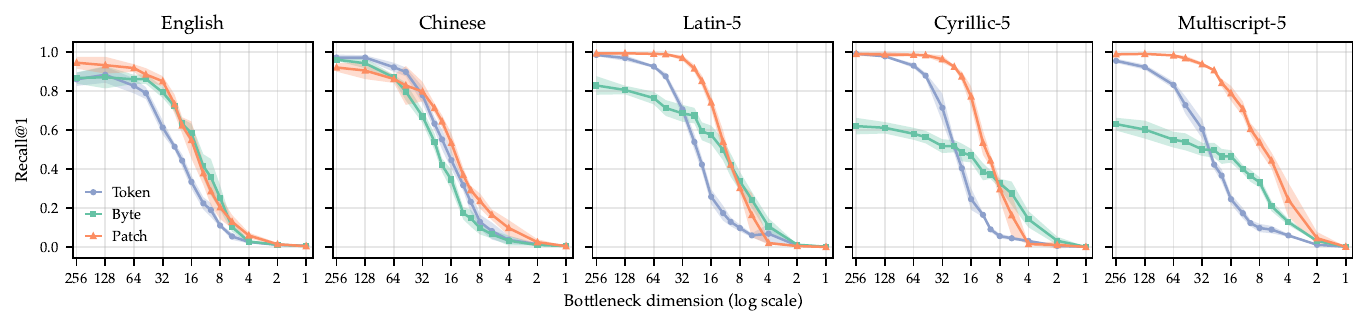}
\vspace{-1.5em}
\caption{%
Form preservation. Test Recall@1 for self-retrieval through a width-$D$ bottleneck, by regime and encoding, averaged over five seeds (shaded band: $\pm{}1$ standard deviation).
Pixels lead wherever capacity is ample.
In the same-script multilingual regimes (Latin-5, Cyrillic-5) bytes overtake below about $D{=}10$.
In Multiscript-5, pixels stay ahead to the smallest latent.
}
\label{fig:form}
\end{figure}

\subsection{Form preservation}
\label{sec:results_form}
The form experiment asks how much of a sentence's surface a single latent vector can hold.
The model encodes one input, compresses it to a latent of width $D$, and is trained to recover its own embedded input from that latent via InfoNCE~\citep{oord2019representationlearningcontrastivepredictive}.
We score recovery by retrieval: a sentence counts as correct if its compressed encoding matches its own input more closely than any other test sentence.
Appendix~\ref{app:task-objectives} provides more details on how retrieval is implemented.
We report test Recall@1 across the sweep (Figure~\ref{fig:form}) but select models on the validation set by mean reciprocal rank (MRR).

Pixels lead wherever capacity is ample.
In the three multilingual regimes a pixel model retrieves almost perfectly down to a width of 48 (Recall@1 $0.97$--$0.99$), and it stays ahead through the mid-range: at $D{=}32$ pixels reach $0.94$--$0.97$, while tokens sit at $0.61$--$0.71$ and bytes at $0.50$--$0.69$.
English and Chinese are closer at the top, but pixels remain the most robust as $D$ falls.

Tokens are competitive only at the widest latents, and they collapse fastest.
In Latin-5, tokens match pixels at $D{=}256$ ($0.98$ against $0.99$) but drop from $0.71$ at $D{=}32$ to $0.26$ at $D{=}16$.
The other multilingual regimes fall just as steeply.
However, bytes behave oppositely.
They are the weakest encoding at the top: at $D{=}256$ byte Recall@1 is $0.83$ in Latin-5 and about $0.62$ in Cyrillic-5 and Multiscript-5, well below pixels, but they degrade slowly.

These opposite slopes cross.
In the two same-script multilingual regimes, bytes overtake both other encodings once the latent is small:
below about $D{=}10$ in Latin-5 and Cyrillic-5, bytes lead, and at $D{=}8$ byte Recall@1 is $0.34$ and $0.33$ against $0.30$ for pixels and $0.10$ and $0.06$ for tokens.
Multiscript-5 is the exception.
There, pixels stay on top to the bottom of the sweep: at $D{=}8$ pixels reach $0.54$ against $0.33$ for bytes, and bytes overtake only tokens, never pixels.

Pixels carry the most distinctive surface, an image of the text, and they are also the shortest sequence to reconstruct (Section~\ref{sec:results_source}), so a generous latent recovers them faithfully.
Compression erodes that surface from the fine detail upward: it keeps coarse, script-level visual features and loses the word-level detail that separates two sentences written in the same script.
In the same-script regimes every distractor shares the script, so the only distinctions left are word-level, which are exactly the details compression destroys.
Therefore, the pixel representations become mutually confusable.
Bytes keep the exact UTF-8 sequence in a 256-symbol alphabet and degrade slowly, so they take the lead once the latent is narrow.
Multiscript-5 differs only in the makeup of the pool and not in the capacity available.
There, four-fifths of the distractors are written in another script, representing a difference coarse enough to survive compression, so they fall away as easy negatives.
Pixel models then only have to separate the same-script sentences.

Sequence length alone does not explain the ordering.
The crossings show that capacity interacts with the structure of each encoding, not only its length.
Tokens make the point from the other side: they are shorter than bytes, yet they collapse first, because each position is part of a large vocabulary and a narrow latent cannot preserve which one.

\begin{figure}[t]
\centering
\includegraphics[width=\linewidth]{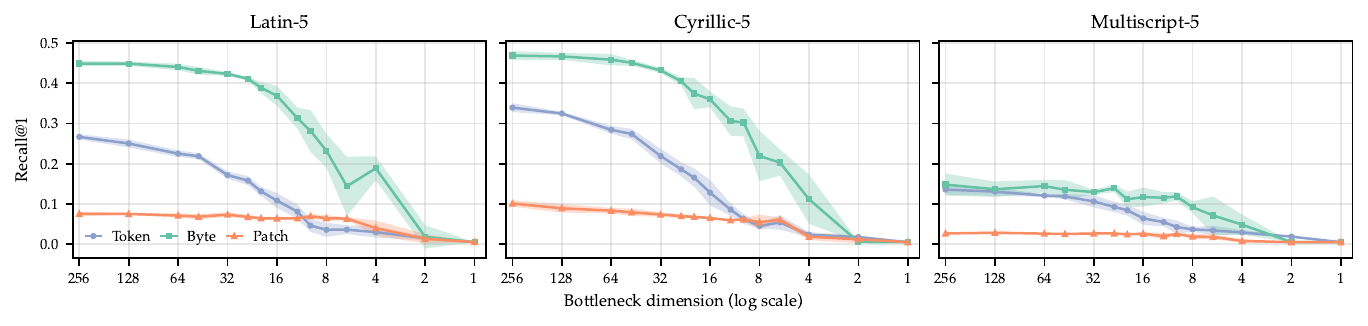}
\vspace{-1.5em}
\caption{%
Cross-lingual retrieval.
Test cross-lingual Recall@1 through a width-$D$ bottleneck, by regime and encoding, averaged over five seeds and over the twenty ordered language pairs (shaded band: $\pm{}1$ standard deviation across seeds).
Bytes lead across widths and regimes.
Pixels are weakest and nearly flat, while tokens are in-between.
All encodings drop sharply in the multiscript regime.
}
\label{fig:semantic}
\end{figure}

\subsection{Cross-lingual retrieval}
\label{sec:results_meaning}

This experiment asks whether translated sentences can still be matched after compression.
It requires no task-specific output head, so the bottleneck latent is the sentence representation.
We train it with a contrastive objective that pulls the five parallel versions of a sentence together and pushes other sentences apart~\citep{khosla2020supervised,oord2019representationlearningcontrastivepredictive}.
For evaluation, each sentence retrieves its translation across every ordered language pair, and we report test Recall@1 averaged over all twenty pairs.
Appendix~\ref{app:task-objectives} provides more details on how retrieval is implemented, and we select models by validation MRR.
Only the three multilingual regimes appear, since cross-lingual retrieval needs at least two languages. 
We note that chance Recall@1 is $1/204{=}0.0049$.

The ranking from the form experiment inverts.
Bytes preserve cross-lingual alignment best, by a wide margin.
At $D{=}256$ a byte model retrieves at $0.45$ in Latin-5 and $0.47$ in Cyrillic-5, against $0.27$ and $0.34$ for tokens and $0.08$ and $0.10$ for pixels (Figure~\ref{fig:semantic}).
Bytes also gain the most from capacity: their Recall@1 falls from $0.45$ to $0.23$ between $D{=}256$ and $D{=}8$ in Latin-5, so most of their advantage is information that a wider latent carries.

Pixels are the weakest encoding at every width, and their curve is nearly flat.
In Latin-5, pixel Recall@1 stays near $0.07$ from $D{=}256$ all the way to $D{=}8$, while bytes more than halve and tokens collapse from $0.27$ to $0.04$.
One hypothesis is that pixels carry little semantic signal to begin with, so the bottleneck has little to keep or lose.
Pixels do better within a script than across one: $0.08$ in Latin-5 and $0.10$ in Cyrillic-5, but only $0.03$ in Multiscript-5 at a third of the same-script level.
Same-script languages share glyphs and cognate spellings, so two aligned sentences look somewhat alike as images.
However, across five unrelated scripts that surface overlap is gone, and pixel retrieval falls with it.

Bytes show the mirror image of the pixel pattern.
Their lead is largest where languages share characters, reaching $0.45$--$0.47$ against tokens' $0.27$--$0.34$.
In Multiscript-5, however, where no two languages share a script, the byte lead over tokens nearly closes ($0.15$ against $0.14$).
Byte-level matching appears to exploit character overlap between languages that share a script.
Over five unrelated scripts, that overlap disappears, so the task is hard for every encoding, with even bytes reaching only $0.15$, a third of their same-script performance.

\begin{figure}[t]
\centering
\includegraphics[width=\linewidth]{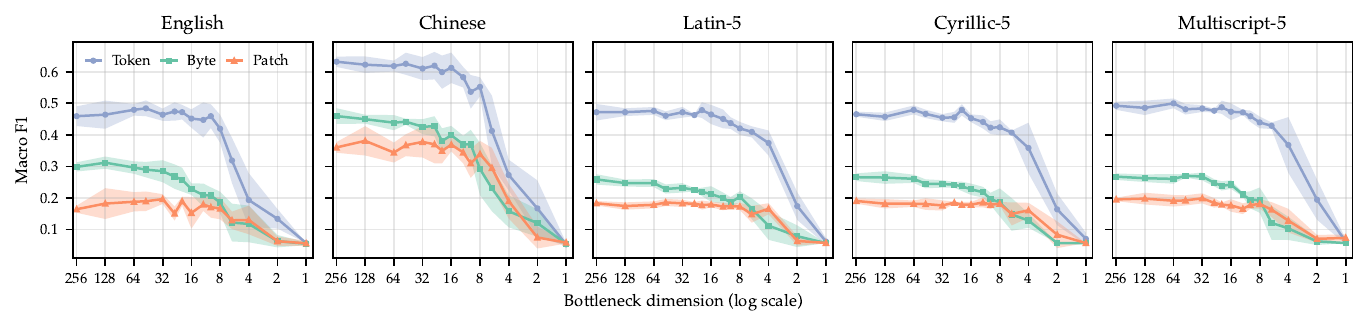}
\vspace{-1.5em}
\caption{%
Topic classification.
Test macro-F1 for SIB-200 topic classification through a width-$D$ bottleneck, by regime and encoding, averaged over five seeds (shaded band: $\pm{}1$ standard deviation across seeds).
Tokens dominate every regime and resist compression best.
Pixels are weakest and nearly flat, except in Chinese, where a meaning-bearing script lifts them.
}
\label{fig:predictive}
\end{figure}

\subsection{Topic classification}
\label{sec:results_predictive}

The topic classification experiment asks whether the compressed representation preserves information useful for a downstream semantic decision.
We attach a linear classifier to the bottleneck output and classify each sentence into one of SIB-200's seven topics.
We select models based on highest validation macro-F1-score, and report test set macro-F1.
All five language regimes are considered.
Labels are content-level, so a sentence carries the same topic in every language.
A uniform guess over the seven topics is $0.14$ macro-F1.

Tokens dominate, by the widest margin of any experiment.
At $D{=}256$, tokens reach $0.46$--$0.50$ macro-F1 in four regimes and $0.63$ in Chinese, against $0.26$--$0.30$ for bytes and $0.16$--$0.20$ for pixels (Figure~\ref{fig:predictive}).
Tokens are also the most robust to compression as they hold most of their performance down to about $D{=}8$.
In Latin-5, macro-F1 slips only from $0.47$ to $0.42$ and collapses only at the smallest latents.

Bytes sit in the middle.
Pixels are weakest, and in every regime except Chinese their macro-F1 stays between $0.16$ and $0.20$ across almost the whole sweep, close to the uniform guess floor and nearly flat, mirroring the shape they showed for meaning (Section~\ref{sec:results_meaning}).
We hypothesize that there is little topic signal in the pixel representation to expose.

Chinese is the exception, and it is the best performing regime for pixels.
While tokens remain best at 0.63 macro-F1, followed by bytes at 0.46, pixels double to $0.36$, marking their largest gain across regimes and closing much of the gap to tokens.
The script is the likely reason.
A Chinese character is a meaning-bearing unit, so an image of Chinese text shows semantic units directly, whereas an image of alphabetic text shows letter shapes that carry meaning only in combination.
Where surface form coincides with meaning, pixels can recover it.

The ordering reflects how directly each encoding exposes lexical content for a coarse semantic decision.
Tokens are dense meaning-bearing units, so a few of them signal a topic and a narrow latent can still preserve which.
Bytes spread the same words over many character positions, so the latent has to reassemble them.
Pixels show surface that, outside a logographic script, is only loosely tied to topic.
This also reverses the ordering from the meaning experiment, where bytes beat tokens.
Character-level overlap helped match translations across languages, but it does not help a topic decision, whereas lexical density does.

\begin{figure}[t]
\centering
\includegraphics[width=\linewidth]{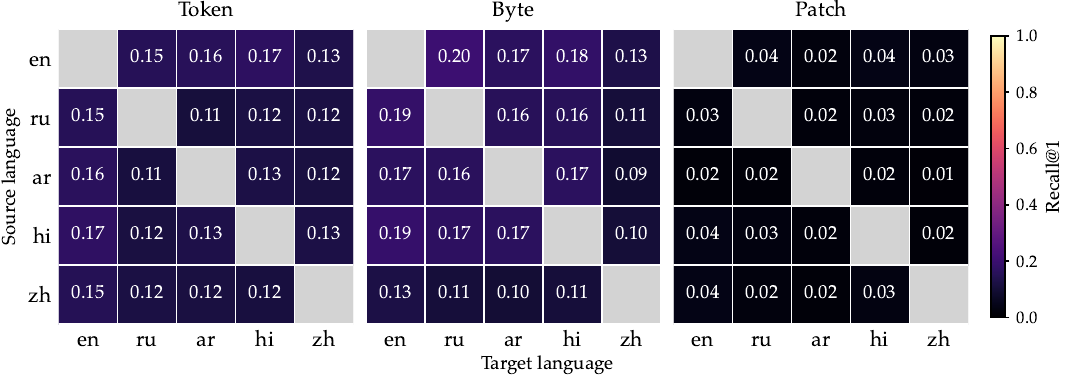}
\caption{%
Per-pair cross-lingual Recall@1 at $D{=}256$, with source language as rows and target language as columns, for each encoding in Multiscript-5 (averaged over five seeds).
Darker cells mark harder retrieval pairs.
Chinese rows and columns are hardest for every encoding.
Pixels are near the $1/204{=}0.0049$ floor throughout.
The same logographic structure that isolates Chinese here is what lets pixels classify it in Section~\ref{sec:results_predictive}.}
\label{fig:kxk_multiscript}
\end{figure}

\subsection{Per-pair alignment, and the Chinese case}
\label{sec:analysis}
Looking at the pair-level breakdown of cross-lingual retrieval shows that the averaged rankings are broad.
Figure~\ref{fig:kxk_multiscript} presents the Multiscript-5 case at $D{=}256$, while Figure~\ref{fig:kxk_all} in Appendix~\ref{app:kxk} provides the full breakdown for all nine multilingual regimes and encodings at $D{=}256$.
Within a regime, the spread across language pairs is smaller than the gap between encodings, with Latin-5 byte Recall@1 running from 0.36 to 0.57 across pairs while pixels do not reach 0.10.
Where languages are close, every encoding does better, especially the surface-bound ones: in Cyrillic-5 the Russian--Ukrainian pair, which is nearly identical in spelling, is the easiest for bytes (0.56) and for pixels (0.14 against a 0.10 average).
Chinese is the outlier.
In Multiscript-5 it is the hardest language to align for every encoding.
Byte Recall@1 on Chinese pairs falls to 0.09--0.13, against 0.16--0.20 among the other four, and pixels sit between 0.01 and 0.04.
A logographic script shares no characters, no subwords, and no visual form with the alphabetic languages, so every encoding struggles to align it.

Yet, the same property reverses sign for topic classification.
Monolingual Chinese is the one regime where pixels classify well (0.36, Section~\ref{sec:results_predictive}).
Chinese characters often correspond to morphemes, so an image of Chinese text exposes units that are more directly tied to lexical meaning than letter shapes in alphabetic scripts.
The same visible structure that isolates Chinese in cross-lingual retrieval can therefore help pixels preserve topic information.

\subsection{Accounting for computation}
\label{sec:results_flops}
The frontiers so far hold content and capacity fixed, but they do not show what each encoding costs to train.
We therefore measure the total training FLOPs of each run: the forward and backward cost of one training epoch, including padding, multiplied by the number of epochs until the best validation score was reached.
Removing all padding from this accounting lowers every total but does not change ordering (Appendix~\ref{app:flops}).

\begin{wraptable}{r}{0.37\linewidth}
\vspace{-1em}
\centering
\small
\begin{tabular}{lrr}
\toprule
Regime & Byte & Patch \\
\midrule
English       & $3.01\times$ & $0.69\times$ \\
Chinese       & $1.50\times$ & $0.29\times$ \\
Latin-5       & $3.00\times$ & $0.66\times$ \\
Cyrillic-5    & $3.78\times$ & $0.55\times$ \\
Multiscript-5 & $4.86\times$ & $0.45\times$ \\
\bottomrule
\end{tabular}
\vspace{-1em}
\caption{Training FLOPs per input, relative to tokens, at $D{=}256$.}
\label{tab:flop_costs}
\end{wraptable}
Total FLOPs are nearly independent of the bottleneck width: sweeping $D$ from 1 to 256 changes the cost of a run by about 4\%, because the shared encoder dominates and the bottleneck is small.
The cost of a run is therefore set by two factors that are independent of capacity: how many positions the input has, and how many epochs the model trains.
The first factor follows the source rates of Section~\ref{sec:results_source}.
Table~\ref{tab:flop_costs} shows the cost of one input relative to tokens: bytes are the most expensive encoding in every regime except Chinese, and patches the cheapest.
The second factor depends on the task.
For form preservation, token models converge in roughly half the epochs of byte and pixel models.
For cross-lingual retrieval the order reverses: byte models converge fastest, on average 40 epochs against 57 for tokens, which offsets part of their higher cost per input.
For topic classification all three encodings converge at similar speed.

\begin{figure}[t]
\centering
\includegraphics[width=\linewidth]{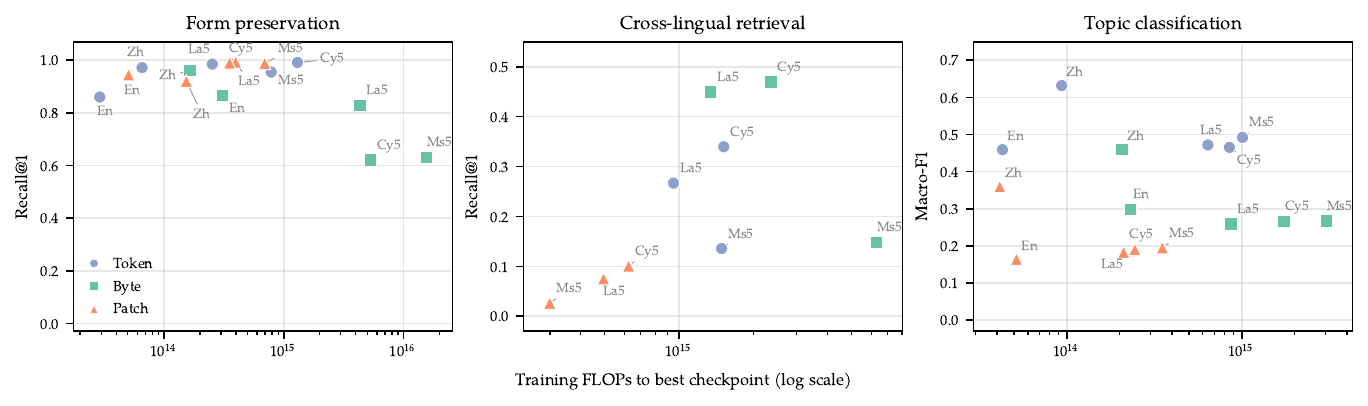}
\caption{%
Test utility against total training FLOPs at $D{=}256$, by task and encoding, averaged over five seeds.
Each point is one language regime (En, Zh, La5, Cy5, Ms5).
Pixels preserve form best and are among the cheapest to train. Tokens classify best at fewer FLOPs than bytes, and the cross-lingual retrieval advantage of bytes costs 1.4--1.6$\times$ the token FLOPs within a script but $4.2\times$ across scripts.
}
\label{fig:flops}
\end{figure}

Figure~\ref{fig:flops} reports test utility against total training FLOPs at $D{=}256$.
For form preservation, accounting for compute strengthens the pixel result.
Pixels reach the best Recall@1 in every multilingual regime and are among the cheapest models to train, while bytes need 11--22$\times$ the FLOPs of pixels and still score lower (0.62--0.83 against 0.99).
The byte overtake at narrow latents (Section~\ref{sec:results_form}) is similarly expensive: at $D{=}8$ in Latin-5, bytes lead pixels by 0.34 to 0.30, but need 4.7$\times$ the FLOPs.
For topic classification, tokens are the most accurate and the cheapest at the top of the utility range.
Bytes reach a lower macro-F1 at a higher cost in every regime, in Latin-5 0.26 against tokens' 0.47 at 1.4$\times$ their FLOPs, so bytes are never the preferred encoding for this task at any budget.
Pixels stay cheapest, and in Chinese they reach over half of token utility (0.36 against 0.63) at less than half of token compute.

Cross-lingual retrieval is the one task where the additional byte compute yields utility that no other encoding reaches.
In the same-script regimes, bytes reach a Recall@1 of 0.45--0.47 against at most 0.27--0.34 for tokens, and the faster convergence of bytes keeps the additional cost at only 1.4--1.6$\times$ the token FLOPs.
In Multiscript-5, the regime without character overlap between languages, this advantage disappears.
Bytes cost 4.2$\times$ the FLOPs of tokens for a gain of only 0.01 Recall@1 (0.15 against 0.14).

\section{Conclusion}
We compared subword tokens, raw bytes, and rendered pixels under matched linguistic content and matched latent capacity.
This control separates three quantities that standard comparisons often collapse: how many input positions an encoding creates, how much information the model is allowed to keep, and which task-relevant information survives.
Across thirteen languages and five scripts, these quantities behave differently.
Source rate varies sharply with language, script, and tokenizer composition, so fixed-length multilingual training does not expose equal content to all languages.

The rate-utility frontiers show that no encoding is best in general.
Pixels preserve surface form most compactly.
Bytes preserve cross-lingual alignment most reliably.
Tokens preserve topic information best.
These differences are not explained by source rate alone.
Short inputs can lose useful meaning, and long inputs can contain structure that compresses well.

The practical takeaway is therefore conditional.
For visual text, OCR-like processing, and other settings where the surface matters, pixels are a strong and computationally efficient interface.
For multilingual retrieval and sentence alignment, bytes deserve stronger consideration, especially when languages share a script.
For encoder-style prediction, tokens remain the strongest default in our experiments.
Input encoding should therefore be chosen alongside the model architecture and training objective rather than treated as a fixed preprocessing step.
Its effectiveness depends on the utility being optimized, the languages being represented, and the capacity available to the system.

\section*{Acknowledgments}
IZ and DE have been supported by the European Union’s Horizon 2020 research and innovation program under grant agreement No. 101135671 (TrustLLM).
This work was supported by a research grant (VIL53122) from VILLUM FONDEN.

\bibliography{colm2026_conference}
\bibliographystyle{colm2026_conference}

\newpage
\appendix

\section{Limitations and future work}
\label{app:limitations-and-future-work}

Our comparison is controlled by design.
This control also limits what the experiments can show.
SIB-200~\citep{adelani-etal-2024-sib} gives us parallel sentences across languages, which is essential for matching content, but it is small compared to open-domain pretraining corpora.
As a result, the token models are not strongly punished for large vocabularies.
The dataset does not fully expose long-tail morphology, domain terms, spelling noise, code switching, or new scripts.
These are settings where bytes and pixels may have a larger advantage than we observe here.
Future work could repeat the comparison on larger controlled parallel corpora.
Resources such as Europarl~\citep{koehn-2005-europarl} or the United Nations Parallel Corpus~\citep{ziemski-etal-2016-united} provide many more aligned sentences than SIB-200, which would make the token setting more realistic and better expose morphology, rare words, and domain-specific vocabulary.

We measure training FLOPs, but not memory or wall-clock time.
Byte activation memory grows with sequence length, and our sentence-length inputs keep the quadratic part of attention~\citep{vaswani2017attention} small.
At document-length contexts, the byte costs of Section~\ref{sec:results_flops} would therefore grow faster than linearly, so our totals understate the byte disadvantage at scale.

Our byte encoder follows the direct byte-processing setup used by ByT5-style models~\citep{xue-etal-2022-byt5}.
It attends over the full byte sequence.
This is a simple and useful baseline, but it is not the only possible byte interface.
Hierarchical pooling~\citep{yu2023megabyte,egli2025multiscale,neitemeier2025hierarchical}, local attention~\citep{beltagy2020longformerlongdocumenttransformer,zaheer2020bigbird}, learned downsampling~\citep{clark-etal-2022-canine,tay2022charformer}, or byte-level compression~\citep{pagnoni-etal-2025-byte} could reduce the cost before the bottleneck.
Such designs may change the tradeoff between source rate and utility.

Our models are encoder-based and are trained for fixed-representation utilities.
They do not evaluate autoregressive generation.
The topic classification task (Section~\ref{sec:results_predictive}) is related to masked or encoder-style inference, where a representation supports a prediction head.
It is only suggestive for decoder-only language modeling, where the output interface, decoding process, and long-context behavior introduce additional constraints.

The pixel experiments also depend on a particular rendering and patching pipeline.
Pixels preserve form well, but they preserve cross-lingual meaning and topic information poorly in our setup (Section~\ref{sec:results}).
It points to a useful open problem: pixel representations have attractive source-rate properties (Section~\ref{sec:results_source}), especially for dense scripts and multiscript settings, but they currently spend much of their capacity on surface detail (Section~\ref{sec:results_form}).
Making them more semantic without losing their compact input interface could improve the compute and multilingual tradeoffs of visual text encoders.

Finally, our experiments use controlled models trained from scratch.
This makes the comparison clean, but it does not replace large-scale pretraining.
Scale, data mixture, tokenizer training, and hardware optimality of certain architectures and objectives may shift the frontiers.

\newpage
\section{Per-pair cross-lingual retrieval}
\label{app:kxk}

\begin{figure}[h]
\centering
\includegraphics[width=\linewidth]{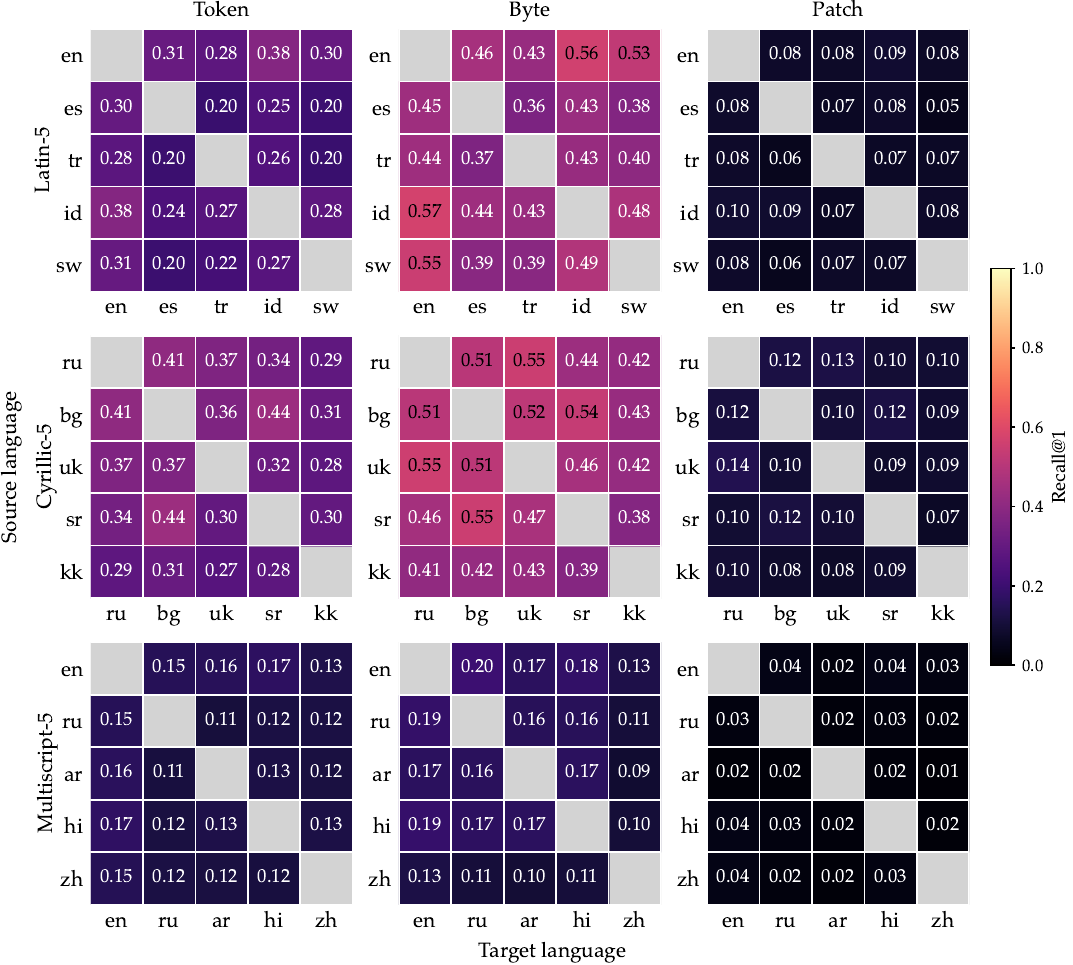}
\caption{%
Per-pair cross-lingual Recall@1 at $D{=}256$ for every multilingual regime and encoding, averaged over five seeds.
Rows are regimes (Latin-5, Cyrillic-5, Multiscript-5) and columns are encodings (token, byte, pixel).
Within each panel, cell $(A,B)$ is Recall@1 for retrieving the language-$B$ translation of a language-$A$ sentence, and darker cells are harder pairs.
}
\label{fig:kxk_all}
\end{figure}

Section~\ref{sec:results_meaning} reports cross-lingual retrieval averaged over all ordered language pairs.
Figure~\ref{fig:kxk_all} shows the full per-pair breakdown behind that average.
For every multilingual regime and encoding, it shows test Recall@1 at $D{=}256$ for each ordered source-target pair, averaged over five seeds, with darker cells marking harder pairs and the diagonal left empty because a language is never retrieved against itself.
The per-pair patterns summarized in the main text are visible directly across the grid.

\section{Experimental details}
We provide code for all experiments at \href{https://github.com/ziegler-ingo/rate-utility-frontiers}{github.com/ziegler-ingo/rate-utility-frontiers}.

\subsection{Input encodings and padded source rates}
\label{app:encoding-details}
All experiments use fixed padded input lengths for each regime and modality.
For tokens and bytes, the padded length is the maximum number of token or byte positions observed in that regime.
For pixels, the padded length is the maximum number of image patches, rounded up to the next multiple of 32 pixels before patching.

\begin{table}[h]
\centering
\begin{tabular}{llrrr}
\toprule
Regime & Language & Tokens & Bytes & Pixels \\
\midrule
\multirow{1}{*}{English}
  & English & \textbf{122} & \textbf{368} & \textbf{81} \\
\addlinespace
\multirow{1}{*}{Chinese}
  & Chinese & \textbf{195} & \textbf{292} & \textbf{54} \\
\addlinespace
\multirow{5}{*}{Latin-5}
  & English & 133 & 368 & 81 \\
  & Spanish & \textbf{137} & 404 & \textbf{87} \\
  & Turkish & 127 & \textbf{411} & 84 \\
  & Indonesian & 116 & 353 & 80 \\
  & Swahili & 124 & 384 & \textbf{87} \\
\addlinespace
\multirow{5}{*}{Cyrillic-5}
  & Russian & 157 & 649 & 88 \\
  & Bulgarian & 150 & 654 & 90 \\
  & Ukrainian & 146 & 624 & 82 \\
  & Serbian & \textbf{187} & 640 & 86 \\
  & Kazakh & 154 & \textbf{708} & \textbf{99} \\
\addlinespace
\multirow{5}{*}{Multiscript-5}
  & English & 161 & 368 & 81 \\
  & Russian & 167 & 649 & \textbf{88} \\
  & Arabic & 143 & 579 & 61 \\
  & Hindi & 168 & \textbf{983} & 67 \\
  & Chinese & \textbf{202} & 292 & 54 \\
\bottomrule
\end{tabular}
\caption{%
Maximum source lengths per language, grouped by regime.
\textbf{Bold} values are the regime-level maxima.
Token and byte rates are sequence lengths.
Pixel rates are counts of non-overlapping $36\times32$ image patches.
}
\end{table}

Token inputs are encoded with the SentencePiece~\citep{kudo-richardson-2018-sentencepiece,kudo-2018-subword} model associated with the regime.
Byte inputs are UTF-8 byte sequences with vocabulary size 256.
Pixel inputs are rendered as grayscale images with height 36.
We render text with Noto Sans fonts at font size 16, using a custom lightweight rendering engine based on HarfBuzz~\footnote{\url{https://github.com/harfbuzz/harfbuzz}} and FreeType~\footnote{\url{https://github.com/freetype/freetype}}.
We crop each rendered sentence to its non-empty horizontal content region, align the crop to patch boundaries, normalize pixel values with mean 0.09 and standard deviation 0.25, and pad with the normalized background value.
Pixels are then projected into patch embeddings with a non-overlapping convolution~\citep{rust2023pixel} over $36 \times 32$ patches.

\subsection{Model and training details}
\label{app:model-details}

All experiments use the same encoder architecture.
Each input first passes through a modality-specific adapter.
Tokens and bytes use learned embedding tables.
Pixels use a non-overlapping convolutional patch projection~\citep{rust2023pixel}.
The adapted sequence is processed by a 6-layer pre-norm Transformer~\citep{pmlr-v119-xiong20b} encoder with hidden size 256, 4 attention heads, RoPE positional embeddings~\citep{su2024rope}, RMSNorm~\citep{zhang2019rmsnorm}, and dropout~\citep{srivastava2014dropout} 0.1.
Attention~\citep{vaswani2017attention} uses a padding mask, so padded positions are ignored.
The encoder output is compressed by a learned-query bottleneck~\citep{pmlr-v139-jaegle21a} with $L=1$ latent and bottleneck width $D$.
We sweep $D \in {256,128,64,48,32,24,20,16,12,10,8,6,4,2,1}$, so the bottleneck rate is $R=D$ in all reported runs.
The bottleneck attention uses $\max(1, \lfloor D/16 \rfloor)$ heads.

Training hyperparameters are shared across experiments unless noted otherwise.
All models are trained for at most 200 epochs with validation after every epoch.
We use AdamW~\citep{kingma2015adam,loshchilov2018decoupled} with learning rate $10^{-4}$, weight decay 0.01, batch size 32, gradient clipping at 1.0, and a learning-rate schedule with 15 warmup epochs followed by cosine decay to $10^{-6}$.
Runs use five seeds: 9109, 2943, 3127, 7081, and 4517.
We select checkpoints by validation MRR for form and semantic utility, and by validation macro-F1 for predictive utility.
Training stops early when the selection metric does not improve for 30 epochs after a minimum training duration of 30 epochs.
For form preservation and cross-lingual retrieval, we also stop when validation Recall@1 reaches 1.0.
For predictive utility, we also stop when validation macro-F1 reaches 1.0.

Table~\ref{tab:appendix-parameter-counts} reports parameter counts for the largest bottleneck setting, $D=256$ and $L=1$.
Counts include the modality adapter, encoder, bottleneck, and task-specific head or readout.
They change with $D$, but we report only the largest setting for compactness.
The adapter size is shown separately because the encodings differ substantially at the input interface.
Token adapter size depends on the tokenizer vocabulary.
The byte adapter has $256 \times 256 = 65{,}536$ parameters.
The pixel adapter has $256 \times 36 \times 32 = 294{,}912$ parameters.

\begin{table}[h]
\centering
\small
\setlength{\tabcolsep}{5pt}
\begin{tabular}{llr rrr}
\toprule
Task & Regime & Token Vocabulary & Token (M) & Byte (M) & Pixel (M) \\
\midrule
\multirow{5}{*}{Form Preservation}
  & English & 11,653 & 9.4 & 6.5 & 6.8 \\
  & Chinese & 4,265 & 7.6 & 6.5 & 6.8 \\
  & Latin-5 & 32,000 & 14.7 & 6.5 & 6.8 \\
  & Cyrillic-5 & 32,000 & 14.7 & 6.5 & 6.8 \\
  & Multiscript-5 & 32,000 & 14.7 & 6.5 & 6.8 \\
\midrule
\multirow{3}{*}{Cross-Lingual Retrieval}
  & Latin-5 & 32,000 & 13.8 & 5.6 & 5.9 \\
  & Cyrillic-5 & 32,000 & 13.8 & 5.6 & 5.9 \\
  & Multiscript-5 & 32,000 & 13.8 & 5.6 & 5.9 \\
\midrule
\multirow{5}{*}{Topic Classification}
  & English & 11,653 & 8.6 & 5.6 & 5.9 \\
  & Chinese & 4,265 & 6.7 & 5.6 & 5.9 \\
  & Latin-5 & 32,000 & 13.8 & 5.6 & 5.9 \\
  & Cyrillic-5 & 32,000 & 13.8 & 5.6 & 5.9 \\
  & Multiscript-5 & 32,000 & 13.8 & 5.6 & 5.9 \\
\bottomrule
\end{tabular}
\caption{%
Parameter counts for the $D=256$, $L=1$ configuration.
Values are in millions (M) and include adapters, rounded to one decimal.
Token adapters are embedding tables and scale with vocabulary size:
11,653 $\rightarrow$ 3.0M, 4,265 $\rightarrow$ 1.1M, 32,000 $\rightarrow$ 8.2M.
Byte adapter is fixed at 0.07M (65,536 params) and pixel adapter at 0.29M
(294,912 params) across all regimes.
}
\label{tab:appendix-parameter-counts}
\end{table}

\subsection{Task objectives and evaluation}
\label{app:task-objectives}

\paragraph{Form preservation.}
The form task measures whether the bottleneck preserves the input representation itself.
For a sentence $x$, the model first computes the modality adapter sequence and uses this detached sequence as the target.
The encoder (Appendix~\ref{app:model-details}) then processes the same input, compresses it through the bottleneck~\citep{pmlr-v139-jaegle21a}, and reads the latent back into a sequence-shaped representation with learned output queries.
We train with a symmetric InfoNCE~\citep{oord2019representationlearningcontrastivepredictive} loss between the readout and the detached target representation.
At evaluation time, every example in the split is encoded once, and each readout retrieves its matching target from the full split.
This task is therefore a contrastive form-preservation probe.
Models are selected by highest validation MRR because MRR is smoother than Recall@1.
We train with a trainable contrastive temperature initialized at $0.07$~\citep{radford2021clip}.

\paragraph{Cross-lingual retrieval.}
The cross-lingual retrieval task measures cross-lingual alignment under compression.
Each batch contains $B$ parallel content units and $K$ languages.
We flatten the batch into $BK$ encoded sentences and train with a multi-positive~\citep{khosla2020supervised} InfoNCE~\citep{oord2019representationlearningcontrastivepredictive} loss.
For each anchor, the positives are all other languages of the same content unit, and the negatives are all sentences from other content units in the batch.
At evaluation time, we build one embedding bank per language.
For every ordered language pair $A \rightarrow B$, each sentence in language $A$ retrieves its translation from all sentences in language $B$.
We average Recall@1 over all $K(K-1)$ ordered language pairs.
Models are selected by highest validation MRR because MRR is smoother than Recall@1.
We train with a trainable contrastive temperature initialized at $0.07$~\citep{radford2021clip}.

\paragraph{Topic classification.}
The classification task measures whether the bottleneck retains information useful for topic classification.
The model compresses the input through the shared bottleneck and uses the single learned query vector to predict the topic with a linear classifier head.
We train with standard cross-entropy.
We report macro-F1 and select models by highest validation macro-F1.

\section{Training FLOP accounting}
\label{app:flops}

Section~\ref{sec:results_flops} reports the total training FLOPs of each run as the cost of one training epoch multiplied by the number of epochs until the best validation score.
This appendix describes how both factors are obtained and shows that the comparison does not depend on padding.

\paragraph{Measurement.}
We measure the cost of one epoch with PyTorch's \texttt{FlopCounterMode}: we build each model exactly as trained, run one forward and backward pass including the task loss, and count every operation, with a matrix product of shapes $(m,k)$ and $(k,n)$ counted as $2mnk$ FLOPs.
Every input is padded to its regime's maximum length (Appendix~\ref{app:encoding-details}), and attention over padded positions is masked but still computed, so the cost of an epoch is the per-input cost times the number of training sentences.
The embedding lookup of token and byte models is a gather and costs no FLOPs. However, its backward pass is a scatter-add, which we count.
The number of epochs is the epoch of the checkpoint that early stopping selected (Appendix~\ref{app:model-details}).
Validation and test passes are excluded.

\paragraph{Removing padding.}
The padded totals reflect the computation our runs performed, but they count every sentence at the length of the longest sentence in its regime.
This could distort the comparison: in Multiscript-5 every byte sequence is padded to Hindi's maximum of 983 bytes, $4.6\times$ the mean byte length.
We therefore recompute every total with padding removed per-sequence.
Every counted operation is a matrix product whose dimensions are affine in the sequence length, so the per-input cost is a degree-two polynomial in the sequence length, with the quadratic term contributed by attention.
We recover the coefficients of this polynomial from three measured probe lengths and verify them against a fourth measured length, which they match exactly.
Summing the polynomial over the true lengths of all training sentences then gives the cost of an epoch without padding.

\begin{figure}[h!]
\centering
\includegraphics[width=\linewidth]{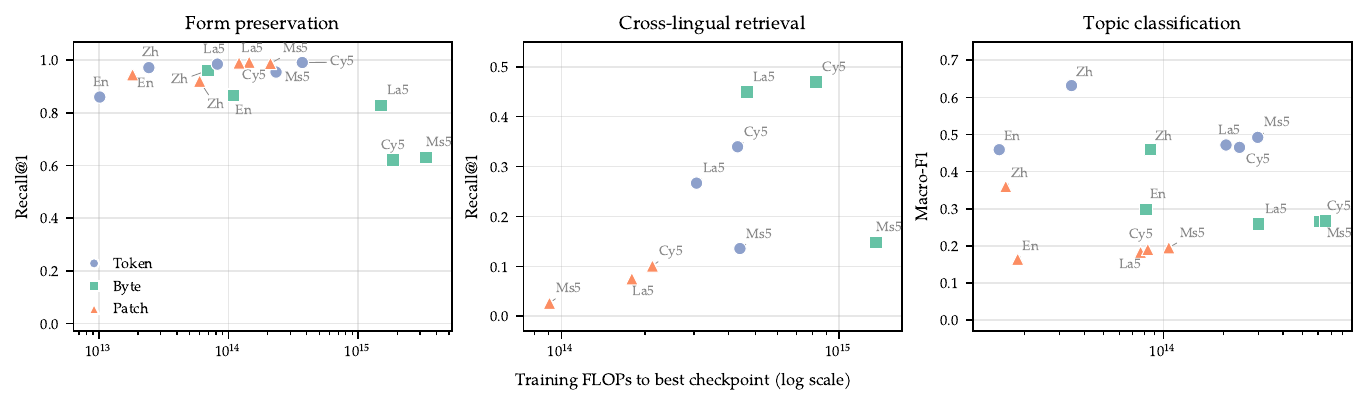}
\caption{%
Recalculation of Figure~\ref{fig:flops} under the zero-padding accounting: test utility against total training FLOPs at $D{=}256$, with every training sentence charged at its own length.
Totals are $2.4$--$4.7\times$ lower than in Figure~\ref{fig:flops}, but conclusions remain unchanged.
}
\label{fig:flops_packed}
\end{figure}

Removing padding lowers every total by 2.4--4.7$\times$ but affects the encodings unevenly.
The per-input cost of bytes relative to tokens falls from 4.86$\times$ to 3.55$\times$ in Multiscript-5, where padding inflated bytes the most, but rises from 3.78$\times$ to 4.66$\times$ in Cyrillic-5, where padding inflated tokens more: the regime's token maximum of 187 is 3.5$\times$ its token mean, driven by a single long Serbian tokenization.
Padding therefore does not systematically favor one encoding.
Figure~\ref{fig:flops_packed} repeats Figure~\ref{fig:flops} under the zero-padding convention.
The conclusions in Section~\ref{sec:results_flops} remain unchanged.

\begin{figure}[h]
\centering
\includegraphics[width=\linewidth]{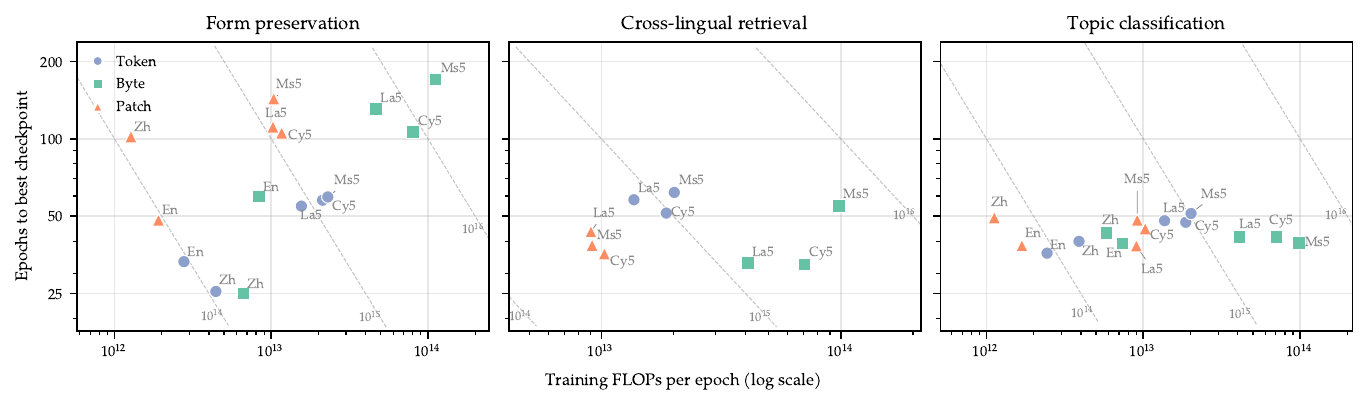}
\caption{%
Decomposition of the training cost into FLOPs per epoch and epochs to the selected checkpoint, by task, regime, and encoding, averaged over all bottleneck widths and five seeds.
Dashed diagonals mark equal total FLOPs.
Bytes are the most expensive encoding per epoch in every multilingual regime, but their position on the vertical axis depends on the task: they are slowest to converge for form preservation but fastest for cross-lingual retrieval.
}
\label{fig:flops_decomp}
\end{figure}

\paragraph{Cost decomposition.}
Figure~\ref{fig:flops_decomp} separates the two factors behind the totals.
Each point places one regime and encoding by its cost per epoch and its number of epochs to the selected checkpoint, averaged over all bottleneck widths and seeds, with diagonal lines marking equal totals.
The horizontal spread repeats the source rates of Section~\ref{sec:results_source}: within each regime, bytes sit rightmost and patches leftmost.
The vertical spread shows the convergence behavior described in Section~\ref{sec:results_flops}: token models converge fastest for form preservation, byte models converge fastest for cross-lingual retrieval, and all three encodings converge at similar speed for topic classification.
Under the zero-padding accounting, the points shift left by the per-input factors above while the epochs stay the same.

\end{document}